\documentclass[twoside,leqno,twocolumn]{article}
\usepackage{ltexpprt}
\usepackage{amssymb}
\usepackage{graphicx}
\usepackage{epsfig,amsmath,subfigure,bm,wrapfig,sidecap,mdwlist} 
\usepackage{url}

\begin{document}

\setlength{\pdfpageheight}{11in}
\setlength{\pdfpagewidth}{8.5in}

\title{\Large Characterizing Driving Styles with Deep Learning}

\author{Weishan Dong\thanks{IBM Research -- China, Beijing, China. \{dongweis, rjyaobj, lcsheng, ytyuanyt\}@cn.ibm.com}
\and Jian Li$^*$\thanks{Nanjing University, China. lionellijian@hotmail.com}
\and Renjie Yao$^*$
\and Changsheng Li$^*$
\and Ting Yuan$^*$
\and Lanjun Wang$^*$\thanks{University of Waterloo, Canada. lanjun.wang@uwaterloo.ca}
}
\date{}

\maketitle


\begin{abstract} \small\baselineskip=9pt
Characterizing driving styles of human drivers using vehicle sensor data, e.g., GPS, is an interesting research problem and an important real-world requirement from automotive industries. A good representation of driving features can be highly valuable for autonomous driving, auto insurance, and many other application scenarios. However, traditional methods mainly rely on handcrafted features, which limit machine learning algorithms to achieve a better performance. In this paper, we propose a novel deep learning solution to this problem, which could be the first attempt of extending deep learning to driving behavior analysis based on GPS data. The proposed approach can effectively extract high level and interpretable features describing complex driving patterns. It also requires significantly less human experience and work. The power of the learned driving style representations are validated through the driver identification problem using a large real dataset.
\end{abstract}

\section{Introduction}
Deep neural networks have been intensively studied in recent years, and many record-breaking progresses have been made on solving computer vision, speech recognition, and natural language processing problems \cite{krizhevsky2012imagenet,abdel2014convolutional,collobert2008unified}. However, so far few attempts have been made on applying deep learning to trajectory data analysis, which is a key research topic in spatiotemporal data analytics, urban computing, intelligent transportation, and Internet-of-Things (IoT) areas.
In this work, we study an important real-world problem of trajectory analysis and propose a deep learning based solution.
The problem comes from the automotive industry, especially the auto insurance and the telematics domain, that is, to characterize the driving styles of car drivers from vehicle sensor data, e.g., GPS (Global Positioning System). 
Because of individual differences, each driver has a signature driving style which is a complex combination of fine-grained driving behaviors and habits. 
Ideally, it should cover the way of accelerating, braking, turning, etc., and their (temporal) combinations given specific driving contexts such as road levels, road shapes, traffic conditions, and even weather.

A good driving style representation is useful in many ways.
For instance, it can be particularly useful in autonomous driving 
that has become a hot topic in both industries and academia.
A better understanding of how human drive a car is helpful to teach machines drive like a human \cite{kuderer2015learning}.
Other examples include to assess drivers' driving risks when correlated with external labels such as claims, accidents, and traffic violations \cite{laurie2011telematics}. In auto insurance businesses (e.g., pay-as-you-drive and pay-how-you-drive), this is a key pricing reference. 
Another common and interesting application is driver identification, i.e., to identify the true driver of anonymized trips \cite{kaggle2015telematics}
, which is useful in scenarios including claim fraud detection, estimating how many drivers share a car for insurance pricing, and the design of intelligent driver assistance systems \cite{lin2014overview}.
When the number of candidate drivers is large (e.g., 1000), it becomes a much harder classification problem than just differentiate safe and unsafe driving behaviors.
If a good driving style representation can solve the driver identification problem, we have reasons to believe that other problems based on driving behavior characterizations can also be solved well.
Therefore, in this paper we take the driver identification problem as a sample task to evaluate the effectiveness of driving style characterization.

The state-of-the-art methods of modeling driving styles are mainly based on handcrafted features \cite{laurie2011telematics,lin2014overview}. 
However, manually defining the driving style by traditional feature engineering is challenging:
(1) It heavily relies on domain knowledge and human experience.
(2) The discriminative power of the features is often unknown before feeding into machine learning algorithms; so a common practice is to enumerate as many features as possible and then apply feature selection, which requires considerable efforts.
(3) The best descriptors of driving patterns may change given different data and contexts, e.g., drivers in China may have different driving patterns from those in US, 
thus a generic model is often hard to obtain.
(4) The feature designs are usually separated from the learning algorithms, which cannot guarantee a best synergia between features and algorithms.
(5) Driving behaviors are typically a sequence of operations, therefore possible combinations of features to define such sequences can be huge. It is hardly to find an optimal driving style representations just by enumerations.

On the other hand, recent advances of deep learning reveal that deep neural network is a promising method of extracting features directly from sensor signal data (e.g., speech). Inspired by this, in this paper, we propose a novel deep learning approach for characterizing driving styles from automotive sensor data, which consists of:
(1) a special design of transforming raw sensor data (typically, GPS) into a form of low-level feature matrices, and
(2) deep neural network architectures based on convolutional neural network (CNN) and recurrent neural network (RNN) to learn high-level driving style features from the low-level feature matrices.
In such a way, complex and higher level discriminative features characterizing how a human driver drives a car can be effectively obtained.

Compared with existing methods, the proposed deep learning approach results in significantly less human work and better accuracy.
To the best of our knowledge, this is the first work of extending deep learning to directly learning driving style representation from automotive sensor data such as GPS.
Experiments on a large real dataset will show that, in terms of identifying the true driver of a trip, the proposed deep learning approach dramatically outperforms the state-of-the-art methods by a large margin. 
If considering additional trip level features (such as trip length, trip shape, etc.) other than just the driving behavior features, performance of traditional machine learning methods can be improved, but is still worse than the proposed deep learning approach that only considers driving style features.
This indicates that deep learning can be a powerful tool for characterizing driving styles.

The remainder of this paper is organized as follows.
Section~\ref{sec:Method} details the proposed deep learning approach.
Section~\ref{sec:Experiments} presents experimental studies on a large real dataset.
Section~\ref{sec:Discussion} discusses a few application related problems.
Section~\ref{sec:Related Work} reviews related work in literature.
Finally, section~\ref{sec:Conclusion} concludes the paper.

\section{Proposed Approach}\label{sec:Method}

The proposed approach consists of two components: data transformation and feature learning by deep networks.
For simplicity, we consider GPS data as the only input.
Nonetheless, as will be discussed later in Section~\ref{sec:Discussion}, our approach can be easily generalized to work with other types of sensor data and rich driving contexts. 

\subsection{Data Transformation: from Geospatial Domain to Movement Statistics Domain}
\label{sec:datatransform}

Deep neural networks have been proved powerful in learning from speech data \cite{graves2013speech,hannun2014deepspeech,abdel2014convolutional}.
GPS sensor data are also kinds of time series having similar characteristics with speech signals, which is the primary motivation for us to develop deep learning methods for driving feature representation learning.
However, a huge gap is that GPS data in its raw format -- a sequence of point geo-locations defined by 2-D coordinates $(x, y)$, each having a timestamp $t$ -- encode spatiotemporal information in an implicit way.
Our empirical studies showed that simply treating raw GPS data as three-dimensional signal inputs for either traditional machine learning or deep learning algorithms just does not work.
A practical way of transforming GPS data sequences (or say, trajectories) into an easier consumable form for deep neural networks needs to be developed.



We define the GPS trajectory as a sequence of tuples \((x,y,t)\), whose length can be varying. 
Inspired by the idea of N-Gram probabilistic model proposed by Brown et al. \cite{brown1992class}, where each word depends only on the last \(n - 1\) words, we utilize the context window concept to make our vision focused on fixed length's trajectory in periods of time.
Specifically, during each window, we may discover potential patterns through observing the behaviors of the driver over different situations. For examples, some drivers may go through a sharp corner quickly while the others would like to slow down, and heavy accelerations can often happen to a group of drivers while the others may never have such aggressive driving behaviors.
In fact, the behaviors are interdependent in time where the length of this period can be \(L_s\). We can roughly say that a current driving behavior depends on what happened in the last \(L_s - 1\) time points. Because of this, there can be more possibility to discover driving patterns if we focus on the trajectory from a windowed perspective.
We can let machine ``understand'' or ``define'' the behaviors in the period of time. To avoid too much information loss, the original trajectory is segmented with a shift of $L_s/2$ so that there is overlap between the neighboring segments.

\begin{figure}[!t]
\centering
\includegraphics[width=0.9\linewidth]{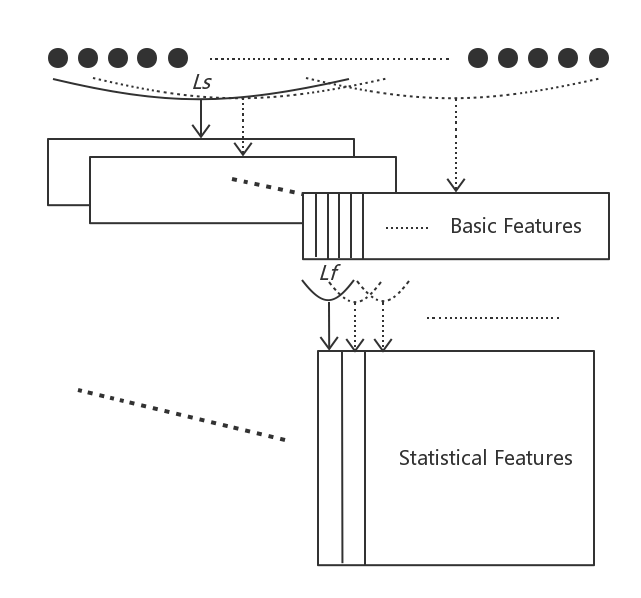}
\caption{Data transformation. First, we generate segments of length \(L_s\) and calculate the basic features from the raw GPS sequence. Then, for each segment, we assign \(L_f\) neighboring points into a frame and generate the statistical features.}
\label{fig1}
\end{figure}

Comparing to the frequency feature map of speech used for deep neural networks \cite{abdel2014convolutional,hannun2014deepspeech}, the next step after we obtain segments from the raw trajectory is to generate feature maps just like the spectrogram of speech which has both axes of frequency and time.
Nonetheless, unlike speech, there is no well-developed method to construct feature maps from trajectories.
We propose to use the following five features to replace the frequency axis of speech data's feature map:
(1) \emph{speed norm},
(2) \emph{difference of speed norm},
(3) \emph{acceleration norm},
(4) \emph{difference of acceleration norm},
and (5) \emph{angular speed}.
We call them as \textbf{basic features} derived from GPS data {at every time point}.
As a result, each segment has a basic feature matrix sized \(5\times L_s\).
Notably, these are all point-wise instantaneous movement features, which are easy to calculate.

To reduce the possible impacts of outliers (which may be generated by small sensor data errors) in such point-wise features, we further derive the statistical information by ``framing'' the segments.
In each segment, we put every \(L_f\) ($L_f < L_s$) neighboring points into a frame with a shift of $L_f/2$, and then calculate the \emph{mean, minimum, maximum, 25\%, 50\%, and 75\% quartiles, and standard deviation} of the basic features in each frame, totally seven statistics.
Such frame level \textbf{statistical features} can be regarded as a more stable representation of the basic features in every time period of length \(L_f\).
The resulting statistical feature matrix ($5\times7=35$ rows representing the driving feature axis and $2\cdot L_s/L_f$ columns representing the time axis) will serve as input to deep neural networks.

In summary, we use a ``large window'' to segment a GPS sequence into fixed-length ``patches'' to model the interdependency among the instantaneous features.
Meanwhile, we employ a ``small window'' to further enframe each segment so as to describe the driving features from the statistical perspectives in a short time period.
Such a double-windowed feature matrix design not only encodes the instantaneous driving behaviors, but also conveys how the patterns change over time.
Importantly, only low-level movement statistics reflecting the driving behaviors are calculated here and no explicit temporal combination is modeled yet.
We expect deep neural networks to learn and extract higher levels of driving style features from such transformed inputs.

\begin{figure}[t]
\centering
    \subfigure[GPS sequence]{\label{fig:datasample}\includegraphics[height=0.4\linewidth]{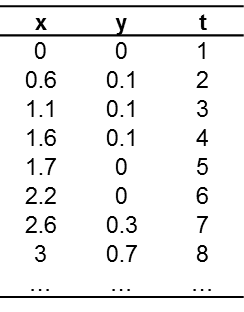}}\quad\quad
    \subfigure[Trajectory view of (a)]{\label{fig:trajsample}\includegraphics[height=0.4\linewidth]{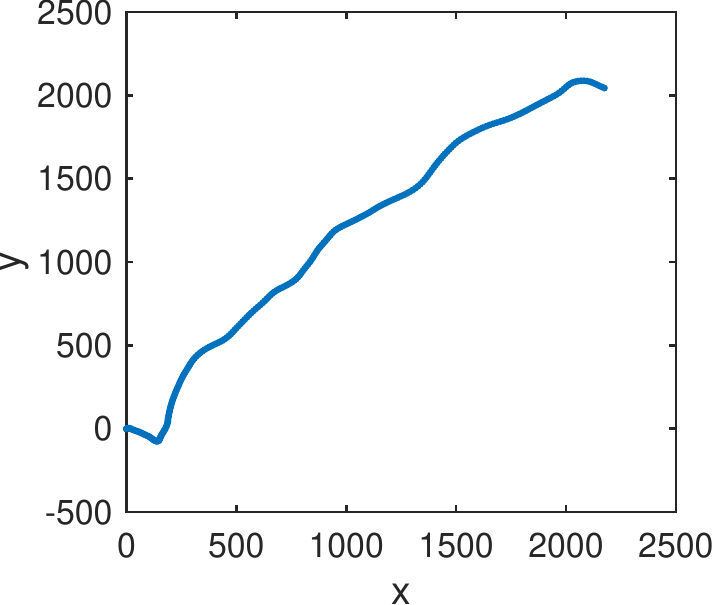}}
    \\
    \subfigure[Statistical feature matrix]{\label{fig:matrixsample}\includegraphics[width=0.9\linewidth]{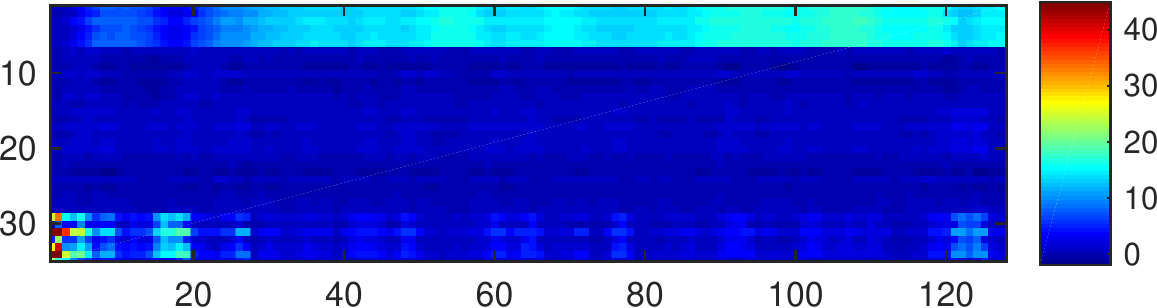}}
\caption{A sample of GPS data transformation. (a) Raw (anonymized) GPS data sequence. (b) Trajectory view of the sequence, showing the first 256 points. (c) Heatmap visualization of the generated statistical feature matrix. 
} \label{example}
\end{figure}

Figure~\ref{fig1} illustrates the proposed data transformation for GPS data sequence from the original geospatial domain to the movement statistics domain.
A long GPS trip thus will be transformed into a number of statistical feature matrices, each corresponds to a segment of the trip.
The label of the original trip (e.g., driver ID) will be assigned to all the segments for supervised learning.
An example of the data transformation is shown in Figure~\ref{example}.
In the example, the GPS data sampling rate is 1Hz, and the location coordinates (in meters) are anonymized with trip origin set to $(0,0)$.
The generated statistical feature matrix sized $35\times 128$ corresponds to a trip segment of $L_s=256s$ 
and $L_f=4s$.
In the feature matrix, the order of rows follows our previous introduction: the speed norm related statistics are in the first seven rows, then follow the difference of speed norm related statistics, and so on. For each basic feature, the order of the seven statistics are also in the same order as in which we listed the statistical features.
We can see that the feature matrix in Figure~\ref{fig:matrixsample} clearly shows the sharp turn at the beginning of the trip (see the last seven rows 29--35, indicating large values of angular speed statistics) and the speed is getting higher over time (see the first seven rows 1--7).


\subsection{Learning with Convolutional Neural Network: Using 1-D Convolution And Pooling}
\label{sec:CNN}

Convolutional neural network (CNN) \cite{lecun1998gradient,krizhevsky2012imagenet} has become popular for image recognitions.
It typically consists of alternating convolution and pooling layers. 
Its two main characteristics, locality and weight sharing, are also beneficial to learning features from time series data such as audio and speech \cite{lee2009unsupervised,abdel2014convolutional}.
We will first employ CNN for learning driving styles from the transformed feature matrix defined in Section~\ref{sec:datatransform}.

\begin{figure}[tb]
\centering
\includegraphics[width=\linewidth]{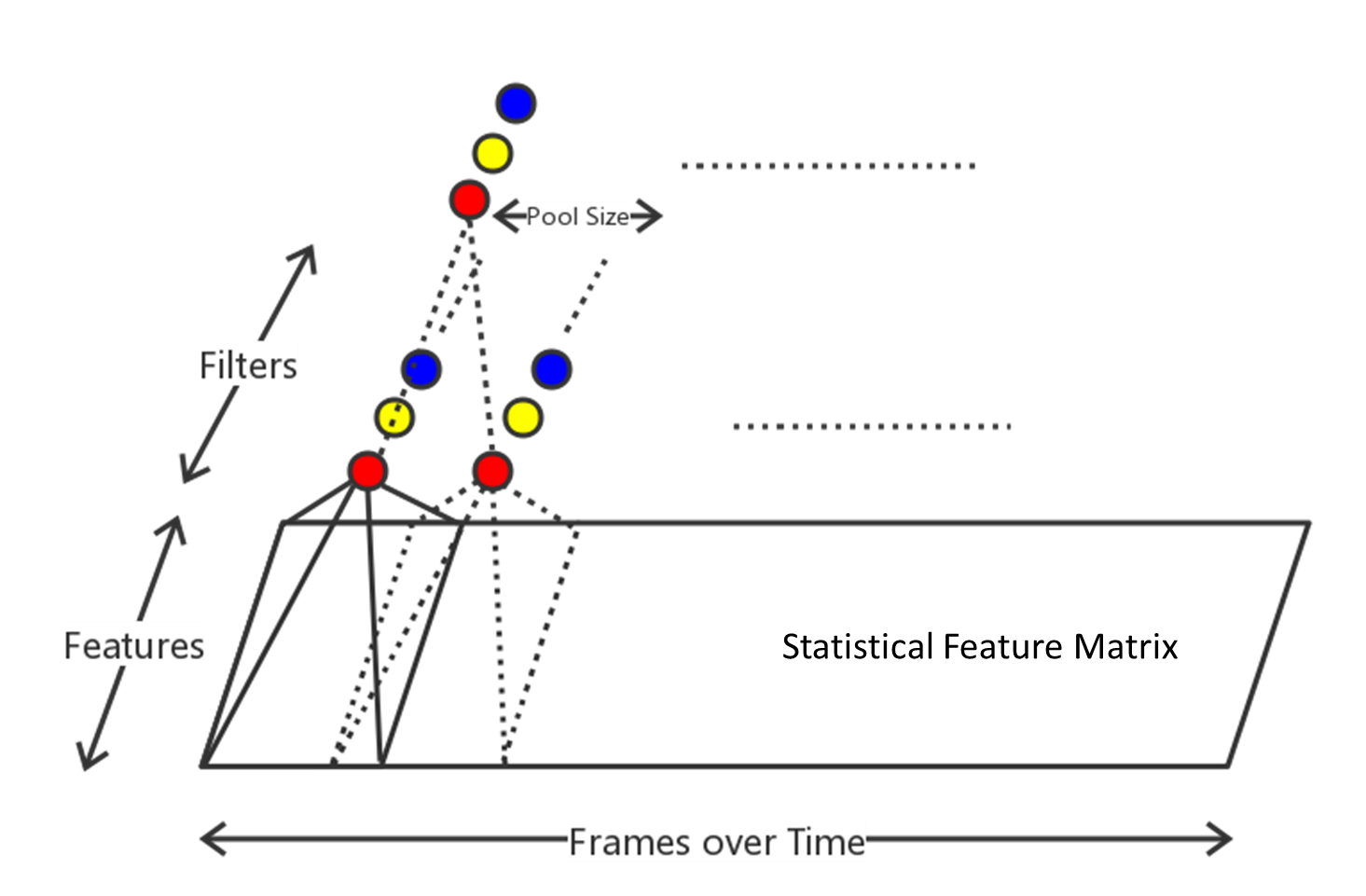}
\caption{1-D convolution and pooling in CNN}
\label{fig2}
\end{figure}

Given the statistical feature matrix data as inputs, we propose to apply 1-D convolutions only over the time axis (columns) because the convolution over driving features has no practical significance.
This is because unlike the frequency axis in speech feature maps where there is local structure from low frequencies to high frequencies in a continuous domain, driving features do not have ordered information and are discrete.
Exchanging the orders of features (i.e., rows) in the statistical feature matrix yields exactly the same input semantically.
In other words, there is no meaningful local structure along the feature axis (rows), thus convolution is not meaningful either.
Similar ideas have been proposed for audio classification \cite{lee2009unsupervised}, where the time-axis convolution helps to learn more effective acoustic features such as phoneme and gender.
In our problem, it maps to learning complex driving behaviors.
Intuitively, the lower convolution layer in the CNN is able to detect fine-grained driving behaviors such as aggressive accelerations, while the higher layer has the ability to find more abstract and semantic level driving patterns.
The pooling layer is another significant component of CNN. Similar to the convolution layers, we propose to apply 1-D max-pooling only over the time axis.
The feature values computed at different time points are pooled and represented by the maximum.
It helps realize translation invariant for driving patterns over time.
An illustration of the 1-D convolution and pooling are shown in Figure~\ref{fig2}.

The CNN architecture we build for the driver identification problem is as follows.
The net has six layers in total. The first three are convolution-pooling layers and the remaining three are fully-connected.
Specifically, assuming the number of frames in each segment is 128, then the first layer filters the \(35\times128\) input data with 32 kernels of \(35\times5\) with a stride of 1 frame.
The second convolutional layer takes as input the pooled output of the first convolutional layer where the pool size is \(1\times2\) and filters it with 64 kernels of size \(1\times3\).
The third convolutional layer also has 64 kernels of size \(1\times3\) connected to the pooled outputs of the second convolutional layer.
Then the fourth and fifth layer are fully connected and have 128 neurons each. Sigmoid activations are applied to the output of every layers.
The last layer is a Softmax, which outputs a distribution over driver IDs, i.e., the class labels of trip segments.

\subsection{Learning with Recurrent Neural Network} \label{sec:RNN}

\begin{figure}[tb]
\centering
\includegraphics[width=\linewidth]{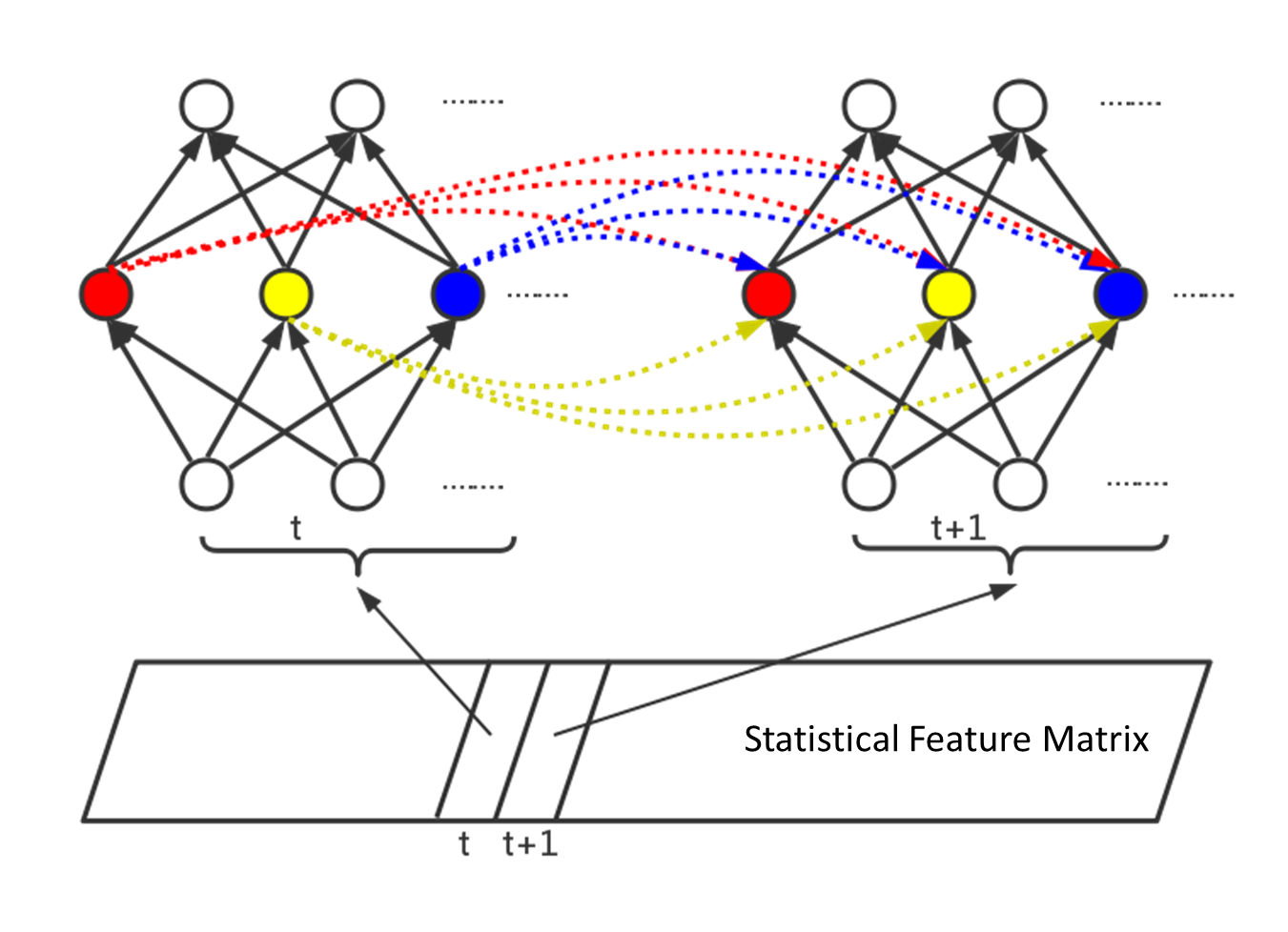}
\vspace{-0.4in}
\caption{RNN training on the input statistical feature matrix}
\label{fig3}
\end{figure}

Recurrent neural network (RNN) is another popular deep neural network that has many variations such as Elman's RNN \cite{elman1990finding}, LSTM \cite{hochreiter1997LSTM} and Bi-directional RNN \cite{schuster1997bidirectional}. It is a kind of feedforward neural network augmented by the inclusion of edges that span adjacent time steps, introducing a notion of time to the model \cite{lipton2015critical}.
Given an input sequence \(s=(s_1,...s_T)\), each neuron in the recurrent hidden layer receives input from the current data point \(s_t\) and also from hidden node values \(h_{t-1}\) in the previous time step:
    \begin{equation}
        h_t = g(W_{hs}s_t+W_{hh}h_{t-1}+b_h)\nonumber
        \label{eq1}
    \end{equation}
where \(W_{hs}\) is the input-hidden weight matrix, \(W_{hh}\) is the matrix of weights between the hidden layer and itself at adjacent time steps, \(b_h\) denote bias vectors, and \(g\) is the hidden layer function.
RNN can be interpreted as an unfolded network across time steps, therefore, it is inherently deep in time. 
Since it is very successful on sequence learning tasks such as speech recognition and machine translation, it is natural to extend RNN to driving style feature learning from GPS data sequences.
In our case, we regard the transformed statistical feature matrix as a sequence of 35-D frames and feed it into RNN.
Figure~\ref{fig3} illustrates how RNN runs on the transformed feature matrix.

As analyzed in \cite{pascanu2012difficulty}, training RNN is difficult due to vanishing and exploding gradients. Researchers have been working on developing optimization techniques and network architectures to solve this problem. Instead of using more sophisticated methods, Le, Jaitly, and Hinton \cite{le2015simple} proposed an RNN architecture composed of rectified linear units (ReLU) and used the identity matrix or its scaled version to initialize the recurrent weight matrix.
Their simple solution is even comparable to a standard implementation of LSTM on certain tasks such as language modeling and speech recognition.
For our driver identification task, we tend to use such a simple yet powerful network, which we denote as IRNN.
The last layer of IRNN, again, is a Softmax appending to the recurrent layer.
We can also construct an IRNN with two stacked recurrent layers, denoted by StackedIRNN.
The output of the first recurrent layer is also a sequence, which connects to the second recurrent layer as input.
The second recurrent layer outputs a hidden-layer feature vector, which is appended by a Softmax as the last layer.
In the next section, we will see that this allows higher level driving feature extraction and leads to better classification performance.

\subsection{Non-deep-learning Baselines}
\label{sec:Baseline}

For comparisons with the proposed deep learning approach, we also propose two non-deep-learning methods as baselines of traditional machine learning approaches for driving style feature learning.
The Gradient Boosting Decision Tree (GBDT) \cite{friedman2001greedy} has been recognized as one of the most powerful machine learning algorithms.
We use it as the first baseline method with feeding the same transformed GPS data as inputs.
But different from CNN and RNN, GBDT does not explicitly model the locality or the time steps in sequence.
Therefore, given the same $35\times 128$ feature matrix as input, GBDT treats it as an unfolded vector of $35\times 128=4480$ features.

As the second baseline representing traditional feature engineering methodology for characterizing driving styles, we also train GBDT on a set of 57 manually defined driving behavior features. We denote it as TripGBDT.
These features were used in a participation to the Kaggle competition on Driver Telematics Analysis \cite{kaggle2015telematics} (the same dataset will be used in experimental studies in the next section) and achieved 0.92 AUC score in detecting false trips in the competition, indicating the effectiveness of these features.

The 57 features consist of global features and local ones.
The global features are trip level statistics, including the mean, min, max, std, and quantiles (25\%, 50\% and 75\%) of speed norm, difference of speed norm, acceleration norm, difference of acceleration norm, and angular speed of a whole trip.
In addition, the following ones are also defined as global features:
time duration of the whole trip,
trip length,
average speed (trip length divided by time duration),
area of the minimal rectangle containing the trip shape, and
lengths of the two edges of the minimal rectangle.
The local features are defined as follows.
We first extract the moving angles (0 to 180 degrees) for each point (based on a window of three consecutive points), and divide them into eight bins [0,10), [10,20), [20, 30), [30,45), [45,60), [60,90), [90,120), and [120,180]. 
In each bin, we also calculate the mean, min, max, std., and quantiles (25\%, 50\% and 75\%) of speed norm, difference of speed norm, acceleration norm, difference of acceleration norm, and angular speed. These features model the correlations between driving behaviors and the road's local shape. 
We first downsample the trip with sampling rates 1, 2, 3, 4, and 5 records per second, and then extract the features on the downsampled trips.
This can be seen as applying a smoothing procedure.
In total, there are 57 features defined.
We can see that not only driving related features are included, but also trip geometry and global statistics are available.
In contrast, in the proposed deep learning approach, only segment level statistics about short-time driving behaviors are calculated. The global information of a trip is completely invisible to the neural networks.
If in such a case, deep learning methods can still outperform TripGBDT, it is convincing to conclude that deep neural networks are more powerful in characterizing driving styles.


\section{Experiments}\label{sec:Experiments}

A major requirement on the data quality for characterizing fine-grained driving behaviors is that the GPS data sampling rate must not be too low. In addition, a regular sampling interval is preferred.
It can be imagined that a low sampling rate can result in too much information loss, especially the instantaneous car movement cannot be estimated accurately.
Our empirical studies revealed that generally when the sampling rate is lower than 0.1Hz (one record per ten seconds), the performance of any approaches can become poor.
We adopt a large public dataset from the Kaggle 2015 competition on Driver Telematics Analysis \cite{kaggle2015telematics} 
for experimental studies.
The dataset contains 547,200 anonymized real trips of 2,736 drivers.
Each driver has 200 driving trips with varying lengths that record the car's position (in meters) every second\footnote{The original problem in the competition was to detect trips that are not driven by a specific driver.
Such ``false'' trips exist in every driver's data, but there is only a small and random number of false trips.
The ground truth of the trips are not available.
Therefore, we regard the driver labels as true labels, and regard the false trips as noise, which does not affect the evaluation much.
}.
We use $L_s=256s,L_f=4s$ to generate the statistical feature matrices from the trip data.
To the best of our knowledge, this dataset is the only publicly available trip dataset having (1) a sufficiently high sampling rate, (2) a regular sampling interval, and (3) a large number of real trips and drivers.
As a result, we are only able to experiment on this one dataset.
We conduct two experiments:
In the first small scale test, we use 50 drivers' data.
In the second large scale test, we use 1,000 drivers' data.
In all the tests, for each driver we randomly select 80\% trips as train data and 20\% as test.
Note that since we segment each trip into fixed-length segments and transform them into the statistical feature matrices, both training and testing are performed on segment data instead of trips.
As for the final evaluation, we care not only the segment-level predictions but also the trip-level predictions.
Once the prediction of each segment of a trip is obtained, the trip-level prediction is calculated through adding up all segments' predictions for a weighted vote.
As the driver identification is a classification problem, we will report segment-level accuracy, trip accuracy, and trip top-5 accuracy in experiments.

\subsection{Candidate Methods And Parameter Settings}
Using the same input data, we train five deep neural networks for comparisons:
\begin{itemize}
  \item CNN: see Section~\ref{sec:CNN}
  \item NoPoolCNN: CNN without pooling layers
  \item IRNN: see Section~\ref{sec:RNN}, with 100 neurons in the recurrent layer
  \item PretrainIRNN: Use the features extracted at the third convolutional layer in the pre-trained CNN as inputs to train an IRNN
  \item StackedIRNN: see Section~\ref{sec:RNN}, with 100 neurons in each recurrent layer
\end{itemize}
In addition, we include GBDT and TripGBDT introduced in Section~\ref{sec:Baseline} in comparisons.

The parameters of algorithms are tuned using the standard 5-fold cross-validation.
We use batch size 128 for training the neural networks.
For CNNs, we use the stochastic gradient descent optimizer with learning rate 0.05, decay 1e-6, and Nesterov momentum 0.9.
For RNNs, we use the RMSProp optimizer with learning rate 1e-6, $\rho$=0.9, and $\epsilon$=1e-6.
For GBDT and TripGBDT, the max tree depth 6 is used in the 50-driver experiments and the stopping iterations leading to the best performance are chosen.
In the 1000-driver experiments, the max tree depth 20 is used instead for TripGBDT.

\subsection{Experiment on 50 Drivers' Data}
The training dataset constructed from the first 50 drivers' trips includes over 35,000 segments, which are greatly augmented compared with the original 8,000 trips.
The best results obtained by each algorithm are summarized in Table~\ref{tab:50driver}.
We can find that IRNN demonstrates strong advantages over the others.
And the obtained accuracies are quite acceptable considering that random guess on trip accuracy should be 2\% (1/50).
Not surprisingly, GBDT performs the worst among all.
The best result (bolded in table) is from StackedIRNN.
Although with a simpler architecture of just one recurrent layer, IRNN still easily beats the remaining candidates.
NoPoolCNN performs worse than CNN, indicating the effectiveness of pooling layers.
PretrainIRNN performs better than CNN, however, it still does not outperform IRNN or StackedIRNN that directly run on the feature matrices.
But it worth to mention that IRNN's training time is much longer than CNN and GBDT.
The more complex StackedIRNN only exhibits small improvement over the single layer IRNN, while the training time cost is nearly doubled.
For TripGBDT, only trip level accuracies are available because there is no segment level training or testing.
Although the trip level global information is provided in addition to the driving related features, TripGBDT cannot perform as good as IRNN and StackedIRNN but better than CNN on this small scale test.
In general, deep neural networks are capable of learning good driving style representations from the transformed feature matrix and can perform better than traditional methods.

       \begin{table}[htb]
       \caption{Results on 50 drivers' data}\label{tab:50driver}
        \centering  
        \small
            \begin{tabular}{lccc}  
            \hline
            Method &Seg (\%) &Trip (\%) &Trip Top-5 (\%)\\
            \hline\hline  
            NoPoolCNN &16.9 &28.3 &56.7\\
            CNN &21.6 &34.9 &63.7\\
            PretrainIRNN &28.2 &44.6 &70.4 \\
            IRNN &34.7 &49.7 &76.9\\
            StackedIRNN &\textbf{34.8} &\textbf{52.3} &\textbf{77.4}\\
            \hline
            GBDT &18.3 &29.1 &55.9\\
            TripGBDT &- &51.2 &74.3\\
            \hline
            \end{tabular}
        \end{table}

\subsection{Experiment on 1000 Drivers' Data}
We further conduct a larger scale test using the first 1000 drivers' trip data. 
We only include CNN, 
StackedIRNN, and TripGBDT in comparison since they are representative methods of their own categories as shown in Table~\ref{tab:50driver}.
The neural network parameters keep unchanged.
The results are reported in Table~\ref{tab:1000driver}.
We can see that deep neural networks exhibit significantly better scalability: the performance does not decrease much considering the problem becomes a harder 1000-class problem (random guess accuracy 0.1\%).
Contrarily, TripGBDT's performance becomes dramatically worse.
This indicates that if the problem becomes harder, the manually defined features are no longer as powerful as the ones learned by the deep neural networks.
Still, StackedIRNN performs the best. But again, it cost a lot more computational time to converge.

        \begin{table}[tb]
        \caption{Results on 1000 drivers' data}\label{tab:1000driver}
        \centering  
        \small
            \begin{tabular}{lccc}  
            \hline
            Method &Seg (\%) &Trip (\%) &Trip Top-5 (\%)\\
            \hline\hline  
            CNN &23.4 &26.7 &46.7\\
            StackedIRNN &\textbf{27.5} &\textbf{40.5} &\textbf{60.4}\\
            \hline
            TripGBDT &- &9.2 &15.8\\
            \hline
            \end{tabular}
        \end{table}

\subsection{Interpretation of Learned Features}

It is interesting to investigate what kind of features are learned by the deep networks.
Here we report that if looking at the second recurrent layer of the StackedIRNN trained from the 1000 drivers' data, among all the training samples, which ones result in the maximum activations on the 100 hidden neurons.
For each neuron, we visualize the training samples (segments) that lead to the top five activations.
By observing the common patterns among the samples, we can analyze what kinds of features have been learned.

\begin{figure*}[ht]
\centering
    \subfigure[]{\label{fig:h21}\includegraphics[width=0.32\linewidth]{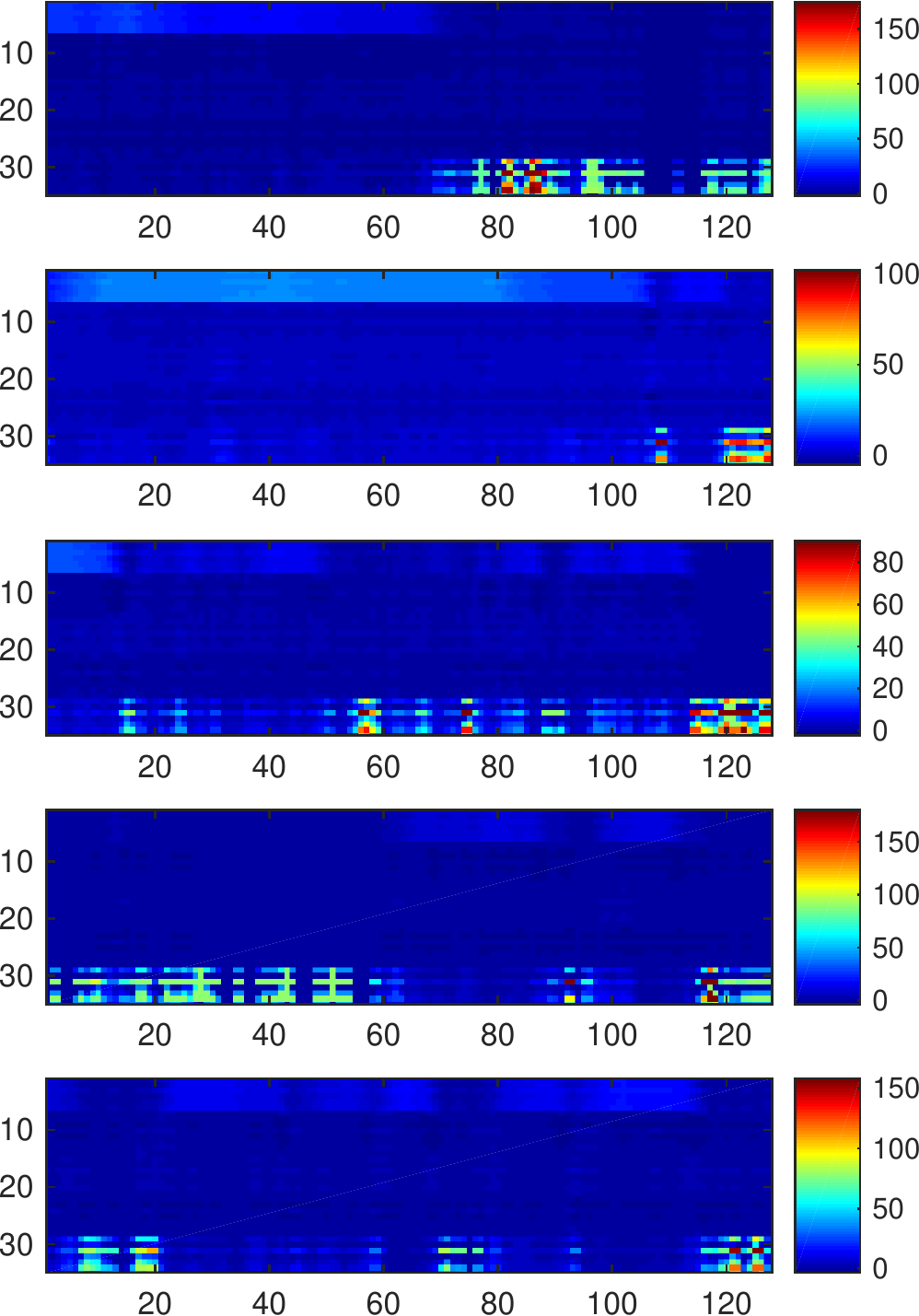}}
    \subfigure[]{\label{fig:h71}\includegraphics[width=0.32\linewidth]{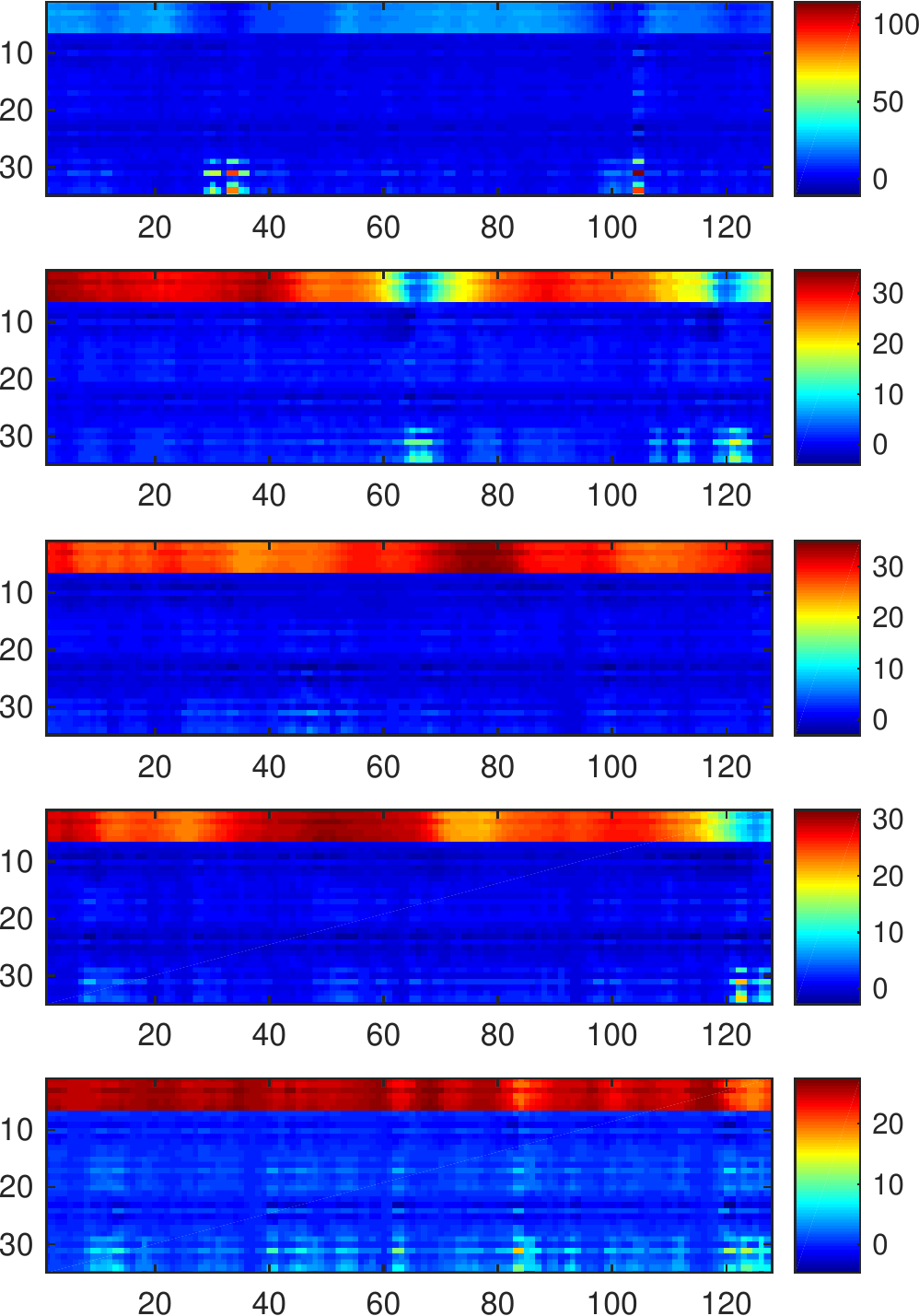}}
    \subfigure[]{\label{fig:h51}\includegraphics[width=0.32\linewidth]{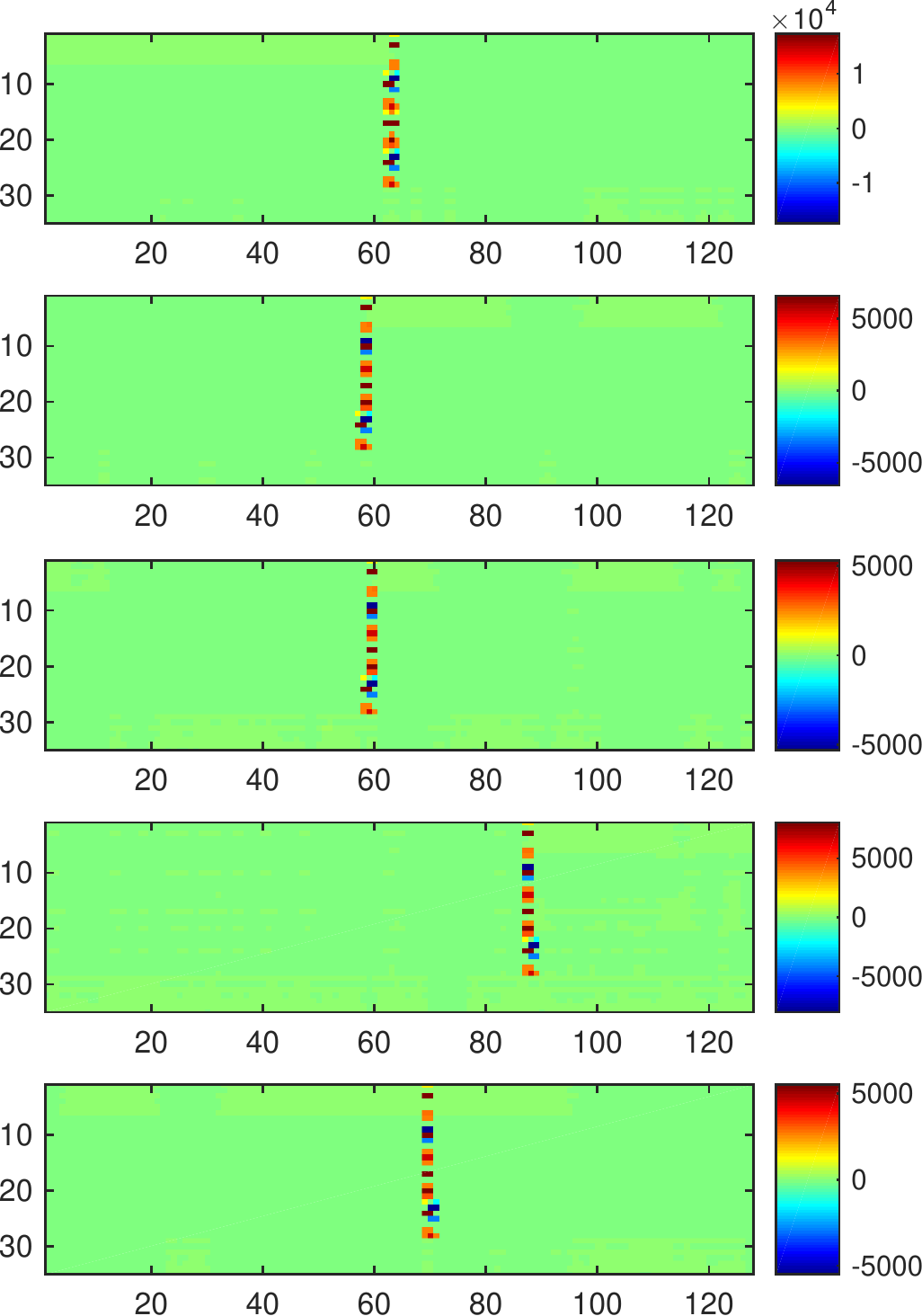}}
\caption{Top five samples making neurons most activated. 
Neuron (a) is activated when it sees the data containing a great number of high angular speed frames and cars slow down around such places. Neuron (b) becomes activated when the input reflect constantly high speed movements along straight roads (angular speeds and accelerations generally keep low but speeds are high). Neuron (c) is the most interesting: it gets activated when the data signals contain a sudden huge jump that may be caused by GPS failures (see the huge magnitudes of speed and acceleration related statistics). } \label{learnedfeature}
\end{figure*}

In Figure~\ref{learnedfeature}, we show the results of three selected neurons.
Each column illustrates the top five training samples that mostly activated a selected neuron.
Interestingly, the neurons seem to have learned driving behaviors such as slowdown at hard turns, 
high speed driving along straight roads, and even those GPS failures that caused sudden huge jumps, i.e., outliers.
These learned features are fairly interpretable, which demonstrates the power of the proposed deep learning approach on driving style feature learning and partially explains its good performance.

\section{Application Related Discussions}\label{sec:Discussion}
In application scenarios requiring real-time prediction, the proposed deep learning approach has a significant advantage over the traditional methods such as TripGBDT that rely on trip level features.
Because only segment level data is needed, given a pre-trained network, the deep learning approach can be used in real-time prediction where segment data can be available online as the car moves.
In contrast, trip level features such as trip length and trip time duration can only be available after the trip ends. This restricts traditional methods such as TripGBDT to be used for online prediction purposes, whereas the deep learning approaches are far more flexible.
It is possible to build an online system based on the proposed deep learning approach, e.g., to predict the driver identity based on the data of a partial trip collected during runtime.
The prediction can be dynamic over time, which is based on the aggregation of predictions on all the so-far collected trip segments.

Privacy is often a key concern in analyzing telematics data. The proposed data transformation in Section~\ref{sec:datatransform} has a merit of not revealing any location or time specific information to the learning phase, even if the GPS trip is not anonymized.
This is because the basic features only describe movements with relative location and time information.
In a real system, if the data transformation can be done, e.g., on the vehicle side, data privacy can be well preserved even if the learning is performed in a centralized manner such as on the cloud side, which requires data to be uploaded for analyses.
This makes it easier to deploy in the real-world.

Driving contexts, e.g., road level, road shape, traffic, and weather, can also influence driving behaviors. Additional car sensor data, such as OBD (On-Board Diagnostic) monitoring the vehicle status, are also helpful to model driving behaviors. Such contextual and sensor data inputs can be further plugged into our framework to enrich the statistical feature matrix. As long as the data are in the format of sequences or time series, similar transformation can be applied so that the calibrated inputs to deep neural networks can encode richer information.
Deep neural networks should be able to discover the discriminative correlations among the feature matrix's elements and learn even better driving style representations.

\section{Related Work}\label{sec:Related Work}

Most existing methods in the literature on driving style modeling rely on a human-defined driving behavior feature set, which consists of handcrafted vehicle movement features derived from sensor data \cite{laurie2011telematics,lin2014overview}.
These features typically work with traditional machine learning methods (supervised classification, unsupervised clustering, or reinforcement learning) to solve problems such as driver identification/classification, driver performance assessment, and human driving style learning \cite{lopez2012driver,Quintero2012,quek2013driver,van2013driver,li2015unsupervised,kuderer2015learning}.
However, as discussed in the introduction, designing the best driving style descriptor is often challenging even for experienced domain experts.
Taking the driver identification problem as examples, the number of classes (i.e., distinct drivers) in the literature is mostly less than ten, indicating the difficulty of developing discriminative feature definitions.
In contrast, our proposed deep learning approach directly work on raw GPS data and automatically learn driving style features, which requires little human work on feature engineering.
Even if the problem size grows to 1000-class as in the experiments, the classification performance is still far better than traditional methods.

There are a lot recent work on deep learning for autonomous driving or Advanced Driver Assistance Systems (ADAS) using camera sensor data as inputs, e.g., \cite{huval2015empirical,chen2015deepdriving,jain2016brain4cars}.
However, the primary purpose of these studies is not to characterize human driver's driving styles.
More importantly, unlike our approach, they are technically about solving computer vision problems rather than learning from GPS records.


\section{Conclusion}\label{sec:Conclusion}

In this paper, we proposed a deep learning approach for characterizing driving styles, which could be the first attempt of extending deep learning to driving style feature learning directly from GPS data.
First, we proposed a data transformation method to construct an easily consumable input form (the statistical feature matrix) from raw GPS sequences for deep learning.
Second, we developed several deep neural network architectures including CNNs with using 1-D convolution and pooling and RNNs and studied their performance on learning a good representation of driving styles from the transformed data inputs.
Taking the driver identification as a sample task, experiments on a large real dataset showed that the proposed deep learning approach significantly outperforms traditional machine learning methods as well as the state-of-the-art feature engineering methods that mostly rely on handcrafted driving behavior features.
Furthermore, the driving style features learned by the deep neural networks were fairly interpretable, explaining the effectiveness of the proposed deep learning approach.
In a word, deep learning can be a powerful tool for learning driving style features from GPS data and many other driving behavior analysis problems.

\bibliographystyle{plain}
\bibliography{DeepLearning}

\end{document}